\pdfoutput=1
\documentclass[11pt,a4paper]{article}
\usepackage[preprint]{neurips_2019}
\usepackage[utf8]{inputenc}
\usepackage{subcaption}
\usepackage{times}
\usepackage{latexsym}
\usepackage{todonotes}
\usepackage{url}
\usepackage{booktabs}
\usepackage{multirow}
\usepackage{makecell}
\usepackage{amsfonts}
\usepackage{amsmath}
\usepackage{float}
\usepackage{diagbox}
\usepackage[graphicx]{realboxes}
\definecolor{light-grey}{RGB}{211,211,211}
\definecolor{celadon}{rgb}{0.67, 0.88, 0.69}
\definecolor{blush}{rgb}{0.87, 0.36, 0.51}
\usepackage{xcolor, colortbl}
\definecolor{OliveGreen}{rgb}{0,0.6,0}

\title{Domain Transfer in Dialogue Systems without Turn-Level Supervision}

\author{Joachim Bingel$^1$, Victor Petrén Bach Hansen$^1$, Ana Valeria Gonzalez$^1$,\\\textbf{Paweł Budzianowski$^2$, Isabelle Augenstein$^1$ and Anders Søgaard$^1$} 
 \\
  $^1$Department of Computer Science, University of Copenhagen, Denmark \\$^2$Department of Engineering, University of Cambridge, UK   \\
  {\tt \{bingel,victor.petren,ana,augenstein,soegaard\}@di.ku.dk}\\
  {\tt pfb30@cam.ac.uk} 
}

\date{}

\begin{document}
\maketitle
\begin{abstract}
Task oriented dialogue systems rely heavily on specialized dialogue state tracking (DST) modules for dynamically predicting user intent throughout the conversation. State-of-the-art DST models are typically trained in a supervised manner from manual annotations at the turn level. However, these annotations are costly to obtain, which makes it difficult to create accurate dialogue systems for new domains. To address these limitations, we propose a method, based on reinforcement learning, for transferring DST models to new domains without turn-level supervision. Across several domains, our experiments show that this method quickly adapts off-the-shelf models to new domains and performs on par with models trained with turn-level supervision. We also show our method can improve models trained using turn-level supervision by subsequent fine-tuning optimization toward dialog-level rewards.
\end{abstract}

\section{Introduction}
Intelligent personal assistants, such as Amazon Alexa, Apple Siri and Google Assistant, are becoming everyday technologies. These assistants can already be used for tasks such as booking a table at your favorite restaurant or the flight for your next vacation. Such dialogue systems potentially allow for smooth interactions with a myriad of online services, but rolling them out to new tasks and domains requires expensive data annotation. In developing goal-oriented dialogue systems, dialogue state tracking (DST) refers to the subtask of incrementally inferring a user's intent as expressed over a sequence of turns. The detected user intent is then used by the dialogue policy in order to decide what action the system should take \citep{henderson2015machine}. For example, in a chatbot-based train reservation system, DST amounts to understanding key information provided by the user as \textit{slot-value pairs}, such as the desired departure and arrival stations, the day and time of travel, among others. With the introduction of the Dialogue State Tracking Challenges \citep{williams2013dialog}, this line of research has received considerable interest. 

State-of-the-art models for dialogue state tracking are typically learned in a fully supervised setting from datasets where slots and values are annotated manually at the turn level \citep{mrkvsic2017neural,zhong2018global,ren2018towards,nouri2018gce}. This allows for high-accuracy models in a select number of domains, where turn-level annotations are available. However, such annotations are cumbersome and costly to obtain, and, in practice, a bottleneck for producing dialogue systems for new domains. 

In this paper, we present an approach to DST that pretrains a model on a source domain for which turn-level annotations exist, then fine-tunes to other target domains for which no turn-level annotation is directly available. In particular, we use standard maximum likelihood training to induce a supervised model for the source domain, and resort to reinforcement learning (RL) from dialog-level signals (e.g., user feedback) for transferring to the target domain, improving target domain performance and potentially saving massive annotation efforts. In addition to this, we also report consistent gains using dialogue-level feedback to further improve supervised models in-domain. 

\begin{figure*}[t]
    \centering
    \includegraphics[width=.9\textwidth]{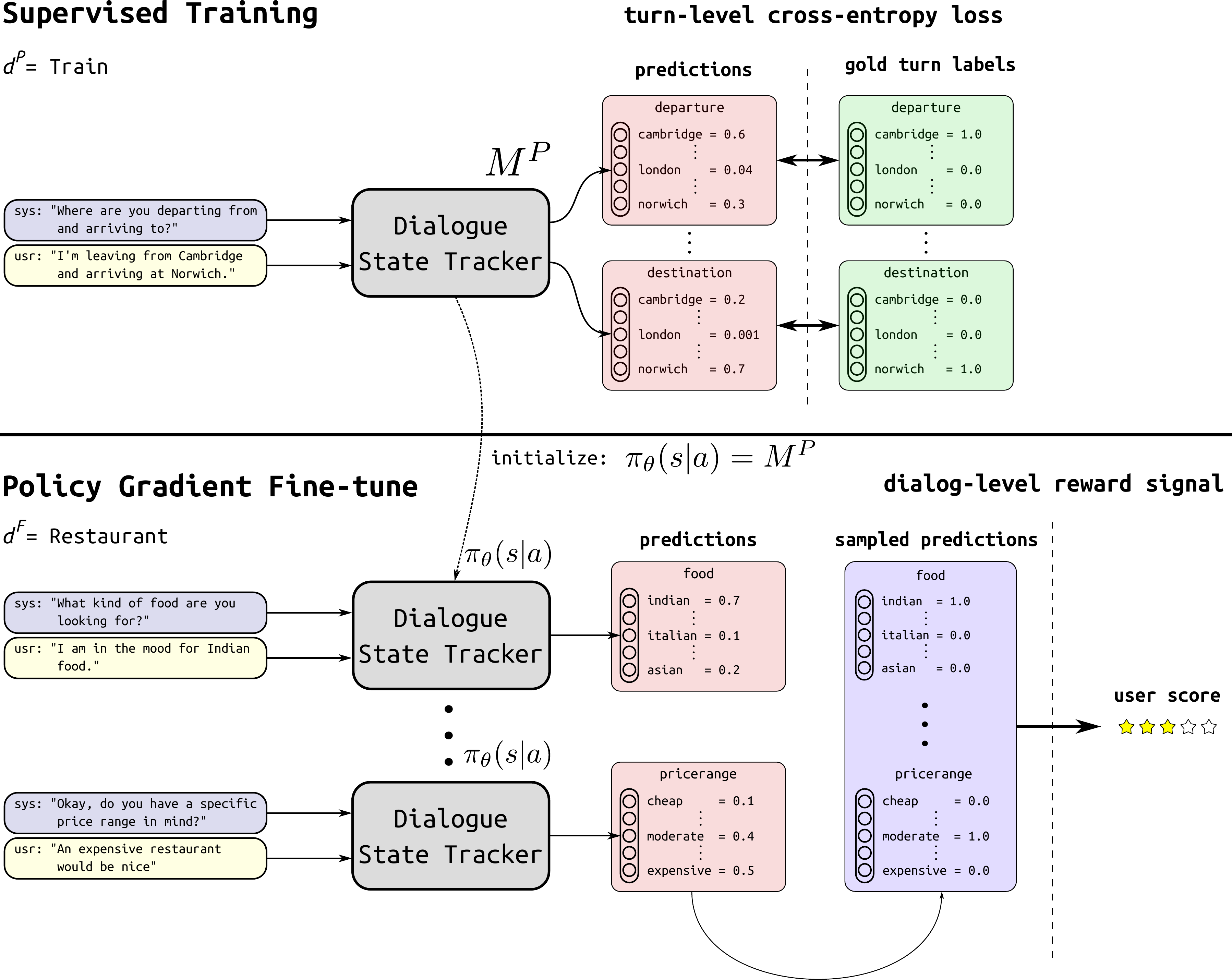}
    \caption{Illustration of our proposed domain transfer dialogue state tracker, using a model $M^P$ trained with turn-level supervision on $d^P$ as a starting point for the fine-tuning policy $\pi_{\theta}(s|a)$ on domain $d^F$.}
    \label{fig:dst}
\end{figure*}

\paragraph{Contributions} To summarize, our contributions are: Relying on \textit{only dialogue-level signals} for target domain fine-tuning, we show that it is possible to transfer between domains in dialogue state tracking using reinforcement learning, gaining a significant increase in performance over baselines trained using source-domain, turn-level annotations.
Second, we show that policy gradient methods can also be used to boost the in-domain accuracy of already converged models trained in the usual supervised manner. 

\section{Baseline Architecture}

\label{sec:dst}
Our proposed model is based on StateNet \citep{ren2018towards}, which uses separate encoders for the two basic inputs that define a turn: the user utterance and the system acts in the previous turn. These inputs are represented as fixed-size vectors that are computed from $n$-gram based word vector averages, then passed through a number of hidden layers and non-linearities. We concatenate these representations, and, for every candidate slot, we compare the result to slot representations, again derived from word vectors and intermediate layers. We update the hidden state of a GRU encoding the dialogue history.and compare this representation to all 
candidate values for a given slot. From this, we compute the probability of slot-value pairs. 
For efficiency reasons, we modify the original StateNet model to only update the GRU that tracks the inner dialogue state after every turn and once all slots are processed within that turn, rather than after every computation of slot values.

Embedding slots and values, and treating them as an input to the model rather than as predefined classes, are important features of StateNet: These features  enable zero-shot learning and make the architecture a natural choice for domain transfer experiments, even if it is not the first to enable zero-shot learning in dialogue state tracking in such a way \citep{zhong2018global,ramadan2018large}. In addition to being well suited for domain transfer, StateNet also produces state-of-the-art results on the DSTC2 and WOZ 2.0 datasets \citep{henderson2014second,mrkvsic2017semantic}.

Training our model is split into two distinct phases. From a pretraining domain $d^P$  for which manual turn-level annotations are available, we learn a model $M^P$, using the available dialogues to train our system until convergence on a held-out development set. Then, for a further domain $d^F \notin D - d^P$, where $D$ is the set of available domains, we use a policy gradient training to fine-tune $M^P$ to the new domain, based on simulated user feedback, corresponding to how 
many goals we met at the end of the conversation.
Figure \ref{fig:dst} presents an overview of this training process.

\paragraph{Pretraining}
In the pretraining phase, we use our implementation of the StateNet model. Just as \citet{ren2018towards}, we focus on predicting the user state and use the information about the system acts contained in the data. During pretraining, we rely on turn level supervision, training models on a single domain and evaluating on a held out set from that same domain. 

\section{Domain Transfer Using Reinforcement Learning}

\label{sec:rl}
\paragraph{Dialogue state tracking with RL}
Given a pretrained model $M^P$ trained on a domain $d^P$, we  fine-tune it on a new domain $d^F$. Since we do not have turn-level annotations for the target domain, we cannot use maximum likelihood training to adapt to $d^F$. This also means that standard domain adaptation methods \citep{Blitzer:ea:06,Daume:Marcu:06,Jiang:Zhai:07} are {\em not}~applicable. Instead, we frame our transfer learning task as a reinforcement learning problem and use policy gradient training. This allows us to use dialogue-level signals as a reward function. Policy gradient training has advantages over value-based RL algorithms, including better convergence properties, ability to learn optimal stochastic policies and effectiveness in high-dimensional action spaces \citep{Sutton:1998}. 
Within this paradigm, the dialogue state tracker can be seen as an \textit{agent} that interact in the \textit{environment} of a dialogue. Throughout the conversation, the DST model tracks the presence of slots in the conversation and assigns a probability distribution over the values, if present. At the end of a dialogue, represented by a state $s$, our model goes through the slots and performs an action, $a$, by sampling a value from the present slot-value probability distribution. It then receives a reward based on how well it predicted slot-value pairs. We illustrate this training regime using dialog-level feedback in the lower half of Figure \ref{fig:dst}.

\paragraph{Dialog-level reward signal} 
In a real-world setting, dynamically obtaining turn-level rewards, for instance from user feedback, is not only costly, but undesirable for the user experience. In contrast, acquiring user feedback at the end of a dialogue, for instance in the form of a 5-star scale, is more feasible and common practice in commercial dialogue systems. 

For practical reasons, we simulate this feedback in our experiments by the success our model achieves in correctly predicting slot-value pairs, assuming that model performance is correlated with user satisfaction. Concretely, we use the Jaccard index between the predicted ($S_P$) and ground-truth ($S_G$) final belief state:


\begin{equation}
    \label{eq:jaccard}
R_{goal}= \frac{|S_G \cap S_P|}{|S_G \cup S_P|}
\end{equation}

\paragraph{Policy Gradient Methods}
We define the policy network $\pi_{\theta}$ as the StateNet network, which is initialized with a pretrained model $M^P$. The weights of the StateNet network are then fine-tuned using stochastic gradient ascent, i.e., in the direction of the gradient of the objective function $\nabla J(\theta)$. The update in the vanilla policy gradient algorithm is:

\begin{equation}
    \label{eq:1}
    \nabla J(\theta) =  \nabla_{\theta} \log \pi_{\theta}(a|s)R_{goal}    
\end{equation}
We update the policy of the network after each iteration, following \citet{Sutton:1998}.

\begin{table*}[t]
  \centering
\begin{tabular}{lcccccc}
\toprule
\multicolumn{1}{l}{Domain} & \multicolumn{1}{c}{Dialogues} & \multicolumn{1}{c}{\makecell{Dialogues\\ with only \\one domain}} & \multicolumn{1}{c}{\makecell{ Turns/ \\Dialogue}} & \multicolumn{1}{c}{Slots} &  \multicolumn{1}{c}{\makecell{Values \\(processed)}} & \multicolumn{1}{c}{\makecell{Split sizes\\ (train-dev-test)}} \\ 
\midrule
\textsc{Taxi}  & 2057   & 435   & 7.66&  4 &  610 & 326-57-52 \\
\textsc{Train}   & 4096 & 345  & 10.26 & 6 & 81 & 282-30-33\\
\textsc{Hotel} & 4197   & 634  & 10.95 & 9 &  187 & 513-56-67  \\
\textsc{Restaurant} & 4692 & 1310  & 8.78 & 6 & 330 & 1199-50-61 \\
\textsc{Attraction} & 3515 & 150 & 7.69 &  2 &  186 & 127-11-12 \\
\bottomrule
\end{tabular}
\caption{Statistics of the MultiWOZ dataset. The reported numbers are from our processed dataset.}
\label{table:multiwoz-stats}
\end{table*}
\paragraph{Variance reduction methods}
Policy gradient methods suffer form certain shortcomings. For instance, they frequently converge to local, instead of global, optima. Furthermore, the evaluation of a policy is inefficient and suffers from high variance \citep{Sutton:1998}. A common way to circumvent the above-mentioned issues is to introduce a baseline model \citep{Weaver:Tao:01}. It is typically initialized as a frozen copy of the pretrained model $M^P$. The baseline models the reward $B_{goal}$ at the end of the dialog. We can then define an \textit{advantage} of an updated model over the initial one as $A_{goal} = R_{goal} - B_{goal}$. In addition to subtracting the baseline, we also add the entropy $\mathcal{H}(\pi_{\theta}(a|s))$ of the policy to the gradient to encourage more exploration  \citep{Williams1991FunctionOU}, in order to counteract the local optima convergence shortcoming. With these modifications to the policy update in Eq. \eqref{eq:1}, we can rewrite the final gradient as:

\vspace{-1.5mm}

\begin{equation}
    \nabla J(\theta)=  \nabla_{\theta} \log \pi_{\theta}(s|a)A_{goal} + \alpha \mathcal{H}(\pi_{\theta}(s|a)),
\end{equation}

where $\alpha$ is a term that control influence of the entropy.


\paragraph{Hill climbing with rollbacks}
Since the policy gradient methods are prone to suffer from performance degradation over time \citep{kakade2002natural}, we employ a rollback method when the policy starts to deviate from the objective. The performance of the model is monitored every few iterations on the development set. If the new model achieves greater rewards than the previously best model, the new model is saved. Contrarily, we roll back to the previous model that performed best and continue from there following other exploration routes if the reward failed to improve for a while. When the policy degrades beyond recovery, the rollback in combination with the slot-value distribution sampling can give a way to a path that leads to greater rewards. We note our hill climbing with rollbacks strategy is an instance of a generalized version of the win-or-learn-fast policy hill climbing framework \citep{Bowling:Veloso:01}. 

\section{Experiments}
\subsection{Data}
We use the MultiWOZ dataset \citep{budzianowski2018multiwoz} which consists of $10,438$ dialogues spanning $7$ domains: \textsc{Attraction}, \textsc{Hospital}, \textsc{Police}, \textsc{Hotel}, \textsc{Restaurant}, \textsc{Taxi} and \textsc{Train}. The dataset contains few dialogues in the \textsc{police} and \textsc{hospital} domains, so we do not include these as the single domain dialogues in these domains did not contain belief state labels. The MultiWOZ dataset consists of natural conversations between a tourist and a clerk from an information center in a touristic city. There are two main types of dialogues. Single-domain dialogues include one domain with a possible booking sub-task. Multi-domain dialogues, on the other hand, include at least two main domains. MultiWOZ is much larger and more complex than other structured dialogue datasets such as WOZ2.0 \citep{mrkvsic2017semantic}, DSTC2 \citep{henderson2014second} and FRAMES \citep{asri2017frames}. In addition, unlike the previous datasets, users can change their intent throughout the conversation, making  state tracking much more difficult. Table \ref{table:multiwoz-stats} presents statistics of domains used in experiments with the distinction between the case when the dialogue consists of only one or more domains. 

\paragraph{Preprocessing MultiWOZ}

The user utterances and system utterances used to trained our models contain tokens that were randomly created during the creation of the data to simulate reference numbers, train IDs, phone numbers, arrival and departure times and post codes. We delexicalize all utterances by replacing these randomly generated values with a special generic token. In addition, we replace the turn label values with this special token and add that to the ontology. As mentioned by \citet{mrkvsic2017neural}, delexicalizing all values is not scalable to large domains as that requires to always have a dictionary holding all possible values. Therefore, we do not delexicalize any other values. 
Since MultiWOZ only contains the current belief state at each turn, we create the labels by registering the changes in the belief state from one turn to the next. The annotators were given instructions on specific goals to follow, however at times they did not follow this goal. This lead to errors in the belief state such as wrong labels or  missing information. These instances also propagate further down to our assigned gold turn labels. Furthermore, while preprocessing the data, we found that there are more values present than reported iin the ontology, therefore the number of values presented here is higher than what is reported in \citet{budzianowski2018multiwoz}.  We release our preprocessed data and preprocessing scripts.\footnote{\label{fn:repo}\url{https://github.com/coastalcph/dialog-rl}}

\begin{table*}[t]
  \centering \small
    \begin{tabular}{l|cc|cc|cc|cc|cc}
    \toprule
    \diagbox{Pretrain}{Finetune} & \multicolumn{2}{c}{\textsc{taxi}} & \multicolumn{2}{c}{\textsc{train}} & \multicolumn{2}{c}{\textsc{hotel}} & \multicolumn{2}{c}{\textsc{restaurant}} & \multicolumn{2}{c}{\textsc{attraction}}\\
                        & \textsc{bl} & \textsc{pg} & \textsc{bl} & \textsc{pg} & \textsc{bl} & \textsc{pg} & \textsc{bl} & \textsc{pg} & \textsc{bl} & \textsc{pg} \\
    \midrule
    \textsc{taxi}       & \cellcolor{blush} 0.35 & \cellcolor{celadon} 0.35 & 0.17 & \textbf{0.27} & 0.04 & \textbf{0.10} & 0.12 & \textbf{0.29} & 0.00 & \textbf{0.11} \\
    \textsc{train}      & 0.13 & 0.13  & \cellcolor{blush} 0.43 & \cellcolor{celadon} 0.43 & 0.07 & \textbf{0.08} & 0.08 & \textbf{0.22} & 0.00 & 0.00 \\
    \textsc{hotel}      & 0.004 & \textbf{0.26} & 0.02 & \textbf{0.19} & \cellcolor{blush} 0.30 & \cellcolor{celadon}  \textbf{0.33} & 0.10 & \textbf{0.19} & 0.06 & \textbf{0.11} \\
    \textsc{restaurant} & 0.04  & \textbf{0.25} & 0.13 & \textbf{0.27} & 0.11 & \textbf{0.13} & \cellcolor{blush} 0.33 & \cellcolor{celadon} \textbf{0.34} & \textbf{0.11} & 0.05 \\
    \textsc{attraction} & 0.00  & \textbf{0.27} & 0.00 & \textbf{0.39} & 0.00 & \textbf{0.08} & 0.05 & \textbf{0.10} & \cellcolor{blush} 0.11 & \cellcolor{celadon} \textbf{0.17} \\
    \midrule
    \textsc{Averages}& 0.04 & \textbf{0.23} & 0.08 & \textbf{0.28} & 0.06 & \textbf{0.10} & 0.09 & \textbf{0.2} & 0.04 & \textbf{0.07} \\
    \bottomrule
    \end{tabular}%
  \caption{Fine-tuning results for our pretrained baseline ({\sc bl}) and the policy gradient (\textsc{pg}). 
  The colored results along the left-to-right downward diagonal are in-domain results, dark red being the supervised results and light green the policy gradient fine-tuned results, and each pair of columns compare baseline and system results for each target domain. The \textsc{Averages} rows presents the average out-of-domain transfer scores for each domain. Note that while the PG method has access to more data, this does not invalidate the comparison, seeing that the additional data is relatively easy to obtain in an applied setting.}
  \label{tab:fine-tune}%
\end{table*}%

\subsection{Implementation Details}
Our pretrained StateNet model is implemented without parameter sharing and is not initialized with single-slot pretraining as in \citet{ren2018towards}. We use the Adam optimizer \citep{KingmaB14} with a learning rate of $10^{-3}$. We use an n-gram utterance representation size of $3$ and $3$ multi-scale receptors per n-gram. The supervised models are trained using a batch size of $16$. The size of the GRUs hidden state is $200$ and the size of the word embeddings is $400$.
In line with recent methods for dialogue state tracking, we use fixed pretrained embeddings and do not update them during the training \citep{mrkvsic2017neural,ren2018towards, zhong2018global}. We use the established data splits for train, development and testing and apply early stopping if the joint goal accuracy has not improved over 20 epochs. 

When fine-tuning with policy gradient, we evaluate on the development set every $5$ batches, saving the model if the reward has increased since last. We use an independent hill climbing patience factor of $15$, reverting back to the previous best model if no improvements were made in that period. We use a batch size of $16$ in our fine-tuning experiments. When applying policy gradient methods in practice, larger batch sizes have shown to lead to more accurate policy updates \citep{papini2017adaptive}, but due to the relatively small training sets we found a batch size of $16$ gave us the best sample efficiency trade-off.
Our implementation uses PyTorch \citep{paszke2017automatic} and is  publicly available.$^1$

\subsection{Experimental Protocol}
\paragraph{Setups} In our experiments, we report a number of different results: 1) Training a DST model $M^P$ with the usual turn-level supervision on the different domains. We only use dialogues which strictly contains the labels of that single domain. We hypothesize that this serves as an upper bound to the performance of the policy gradient fine-tuning. 2) Evaluating the pretrained models as a cross-domain zero-shot baseline. We take a model pretrained on $d^P$ and measure its performance on $d^F$ for all domains in $D - d^P$. This serves as the lower bound for the performance of the policy gradient fine-tuned models. We use this baseline and not a model fine-tuned on $d^F$ with cross entropy training with dialogue level supervision on the final belief state, as we simulate not having gold labels for each slot-value pair, but rather only a scalar rating as the sole signal. 3) Fine-tuning the pretrained model $M^P$ to all other domains with policy gradient as described in Section \ref{sec:rl}. We experiment with domain transfer from $d^P$ to all domains in $D - d^P$ using only the user simulated dialog-level reward using policy gradient. 
4) Lastly, we report the results of fine-tuning a model using policy gradient on the same domain it was pretrained on, $d^P$, after convergence in order to see if the dialog-level reward signal can further improve its performance. We here use the same training and development data as the supervised model was trained on.

\paragraph{Metric} We measure the performance of our models with what we refer to as the \textit{turn level accuracy} metric, which measures the ratio of how many of the gold turn labels are predicted by the DST model at each turn. The reported accuracy is the mean of all turns in the evaluation set.

\section{Results}
In Table \ref{tab:fine-tune} we present the results from our baseline StateNet model and from policy gradient training for the in- and out-of domain scenarios. We also report the average out-of-domain accuracies for each domain, to illustrate how policy gradient training in general performs compared to the baseline. The table show the performance of transferring from each domain to all other domains. From the results we observe that in almost all domain transfer settings, with the exception of \textsc{restaurant} to \textsc{attraction}, we get a consistent increase in performance when applying policy gradient fine-tuning, compared to the zero-shot transfer baselines. In some instances we also see an increase in performance from further fine-tuning a model after turn-level supervision convergence using only the dialogue-level reward feedback. In the case of \textsc{attraction}, we are even able to increase the accuracy by a large margin using in-domain policy gradient fine-tuning. 
On average, we see relative improvements of the accuracy, ranging from $0.03$ to $0.2$, when applying our proposed method of fine-tuning for DST domain transfer.
\section{Analysis}
\begin{figure}
\centering
\begin{minipage}[t]{.475\textwidth} \centering
\includegraphics[width=\linewidth]{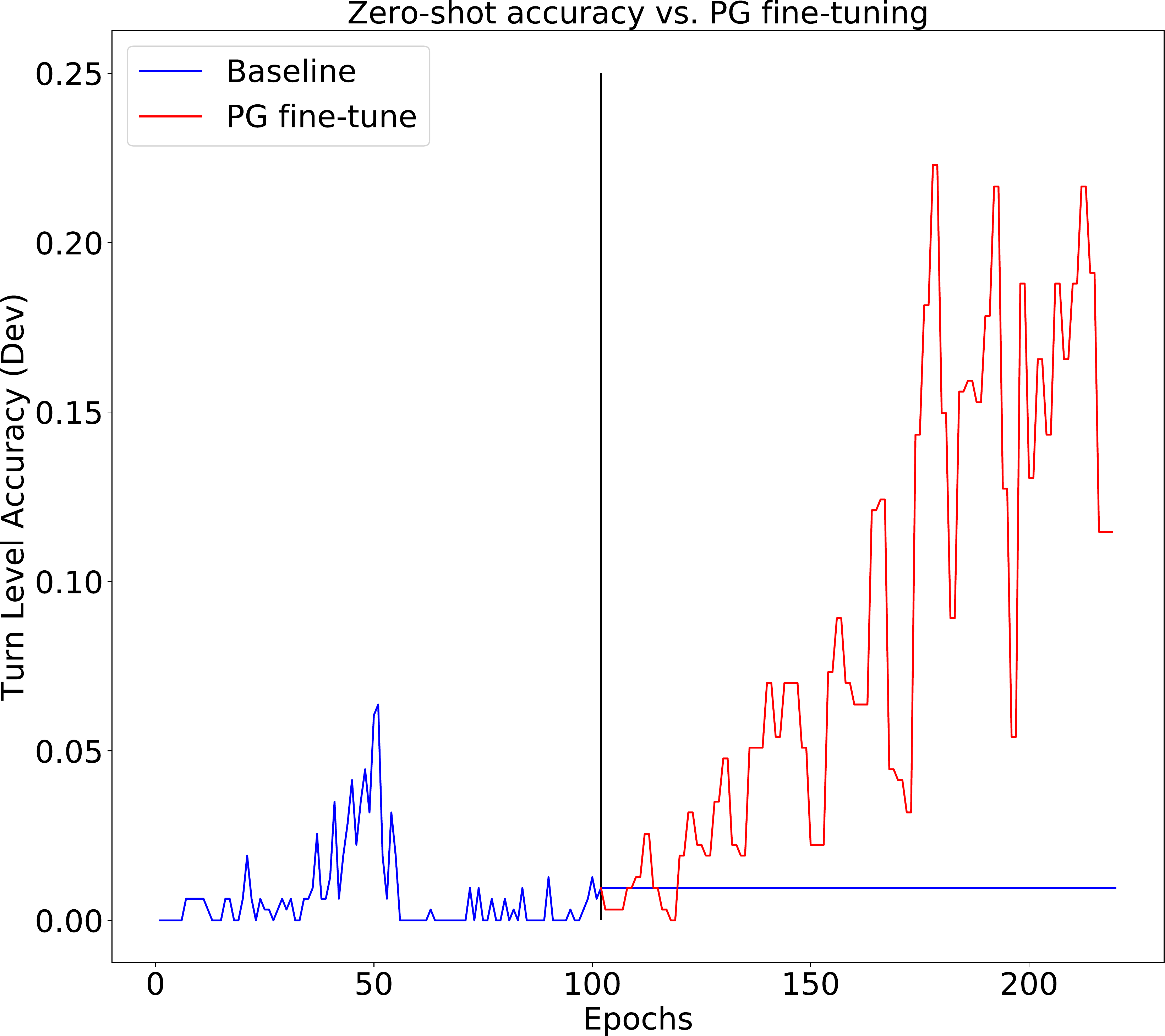}
\caption{The performance of the supervised model trained on the \textsc{hotel} domain while evaluated on the development set of the \textsc{taxi} domain after each epoch until convergence on \textsc{hotel} versus the improvements we get from the policy gradient fine-tuning using the supervised model as starting point.}
\label{fig:ft_improvements}
\end{minipage}%
\hfill
\begin{minipage}[t]{.475\textwidth} \centering
\includegraphics[width=\linewidth]{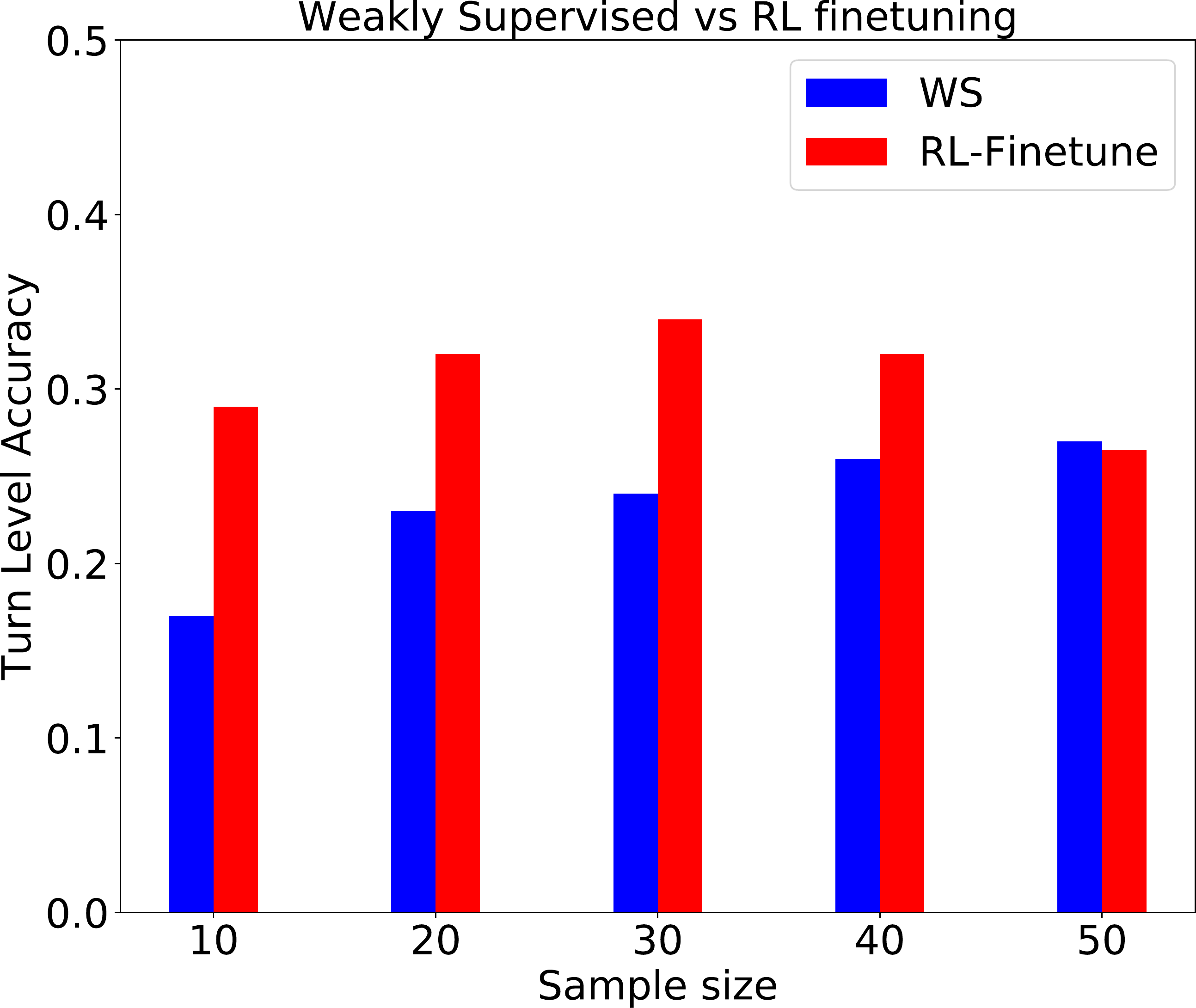}
\caption{The turn level accuracy of our weakly supervised fine-tuning compared to fine-tuning using PG. Performance plateaus after about 50 samples for both methods.}
\label{fig:learning-curve}
\end{minipage}%
\end{figure}

In order to illustrate the effectiveness of doing PG fine-tuning compared to doing zero-shot domain transfer, we plot in Figure \ref{fig:ft_improvements} the results of training a model on the source domain \textsc{hotel} while evaluating, on the development set, its zero-shot accuracy on the target domain \textsc{taxi}, until convergence on the source domain. After convergence we show how the PG fine-tuning uses the pretrained model as a starting point to further improve the accuracy on the target domain using only the dialog-level feedback. Figure \ref{fig:ft_improvements} also illustrates the importance of the hill climbing technique we employ. When the performance starts to deteriorate, it manages to revert back to a reasonable baseline and improve performance from there instead. 
From the blue baseline curve, we also observe that even though the accuracy continuously improves on the source domain, this is not necessarily an indication of the performance on the target domain. On the contrary, performance suddenly starts to deteriorate for the latter when the model overfits to the source domain.

\subsection{Error Analysis}
In general we observe lower scores for both the baseline models and in-domain fine-tuning on the \textsc{attraction} domain. We believe this can be attributed to the fact that it only contains 150 dialogues, leaving very little data for the development and test splits. Coupled with the fact that it has 2 slots and 180 values, the risk of encountering unseen slot-value pairs increases significantly.

In Table \ref{tab:examples} we present a couple of example turns from the test set of the \textsc{restaurant} domain, with the system utterance, user utterance and the predicted slot-value pairs for both the baseline model, which has been trained on the \textsc{hotel} domain, and the PG fine-tuned model. The slot-value pairs in green show correct predictions, whereas pairs in red show incorrect predictions. From the predicted slot-value pairs, we can for example see how the fine-tuned model to a better extent is able to utilize the user and system utterances to correctly predict what price range the user is looking for, even though the baseline correctly predicts the slot presence.



\subsection{Comparisons to Weak Supervision}

We also pose the question of how many annotated dialogues in the target domain are needed before policy gradient fine-tuning with dialogue-level rewards is no longer beneficial, compared to fine-tuning a model trained with turn-level cross entropy.
In order to further investigate this, we use our pretrained model in the \textsc{taxi} domain and further finetune with varying amounts of dialogues i.e. $s\in [10, 20, 30, 40, 50 ]$ using turn level supervision for the \textsc{restaurant} domain. We then fine-tuned each of the models on the \textsc{restaurant} domain using the dialogue-level reward only. The results for these experiments are shown in Figure \ref{fig:learning-curve}. Overall, we find that when we annotate just 10 complete dialogues and then fine-tune our model using reinforcement learning we still see an increase in performance. We observe that as we increase the sample size $s$ for our weakly supervised models, fine-tuning using policy gradient comes with diminishing returns. At around 50 samples, the performance of the weakly supervised baseline reaches the performance of our system, and improvements from reinforcement learning, if any, become significantly smaller.

\section{Related Work}
\paragraph{DST architectures}
The goal of Dialogue State Tracking is to predict the user intent or \textit{belief state} at each turn of the conversation. The range of user goals or, \textit{slots} and \textit{value} pairs, that can possibly be recognized by the system are contained in the domain ontology. DST has for long been a part of spoken dialogue systems, however, before the Dialogue State Tracking challenges \citep{williams2013dialog, henderson2014second} 
 many of the early architectures relied on hand crafted rules \citep{wang2013simple, sun2014generalized, sun2016hybrid}. 
 Later research has proposed RNN models that exploit delexicalized features \citep{henderson2014word,mrkvsic2015multi,rastogi2017scalable} in order to allow the model to perform better and achieve generalization by reducing the amount of labels. Delexicalization requires that all possible mentions of a slot and value are contained in a lexicon which does not become scalable in larger domains. To address this, \citet{mrkvsic2017neural} proposed a neural belief tracker which uses pretrained word embeddings to represent user utterances, system acts and current candidate slot-value pairs and utilizes these as inputs into a neural network.
Recent approaches have proposed sharing parameters across estimators for the slot-value pairs
\citep{zhong2018global,ren2018towards, ramadan2018large,nouri2018gce}. Although  not extensively investigated, this would make the model more scalable as the amount of parameters would not increase while the ontology size grows. In our experiments, we adopt the model by \citet{ren2018towards} as our supervised baseline.

\paragraph{Domain transfer}
A key issue that remains unexplored by many of the existing methods within DST is domain adaptation.
\citet{W13-4068} presented some of the earliest work dealing with multi-domain dialogue state tracking, investigating domain transfer in two dimensions: 1) sharing parameters across slots, 2) sharing parameters across single domain systems. Later research further expanded by using disparate data sources 
in order to train a general multi-domain belief tracker \citep{mrkvsic2015multi}. The tracker is then fine-tuned to a single domain to create a specialized system that has background knowledge across various domains. Furthermore, \citet{rastogi2017scalable} proposed a multi-domain dialogue state tracker that uses a bidirectional GRU to encode utterances from user and system which are then passed in combination with candidate slots and values to a feed-forward network. Unlike our proposed method, they rely on delexicalization of all values. In addition, their GRU shares parameters across domains. 
\citet{ramadan2018large} introduced an approach which leverages the semantic similarities between the user utterances and the terms contained in the ontology. 
In their proposed model, domain tracking is learned jointly with the belief state following \citet{Mrksic:2018acl}.
We want to emphasize that all previous models assume the existence of dialogue data annotated at the turn level in the new domain. In our proposed method, we model a more realistic scenario in which we only have a score of how accurate the system was at the end of the dialogue given the final user goal.



\paragraph{Reinforcement Learning in Dialogue}
In task-oriented dialogues, the reinforcement learning framework has mostly been used to tackle dialogue policy learning \citep{singh2002optimizing,WILLIAMS2007393,Li2009ReinforcementLF, liu2018dialogue}. 
\citet{gasic2013pomdp} proposed a method to expand a domain to include previously unseen slots using Gaussian process POMDP optimization. While they discuss the potential of their model in adapting to new domains, their study does not present results in multi-domain dialogue management.
Recent work has attempted to build end-to-end systems that can learn both user states and dialogue policy using reinforcement learning.  
 \citet{zhao2016} propose an end-to-end dialogue model that uses RL to jointly learn state tracking and dialogue policy. This model augments the output action space with predefined API calls which modify a query hypothesis which can only hold one slot value pair at a time.   \citet{dhingra2017towards} instead show that providing the model with the posterior distribution of the user goal over a knowledge base, and integrating that with RL, leads to higher task success rate and reward.
 In contrast to our work, \citet{gavsic2017dialogue} have tackled the problem of domain adaptation using RL to learn generic policies and derive domain specific policies. In a similar study, \citet{chen2018policy} approach the problem of domain adaptation by introducing slot-dependent and slot-independent agents. 
Our approach differs from the previously presented models in several ways: a) we track the user state using RL, however, we do not learn generic and specific policies ; b) we use RL to adapt models across many domains and a large number of \textit{slot,value} pairs; and c) we assume that a reward is only known for target domain dialogues at the end of each dialogue.   

\section{Conclusion}
This paper tackles the challenge of transferring dialogue state tracking models across domains without having target-domain supervision at the turn level; that is, without manual annotations, which are costly to obtain. Our setup is motivated by the fact that in a practical setting it is much more feasible to obtain dialogue level signals such as user satisfaction. We introduce a transfer learning method to address this, using supervised learning to learn a base model and then using reinforcement learning for fine-tuning using our dialogue level reward. Our results show consistent improvements over domain transfer baselines without fine-tuning, at times showing similar performance to in-domain models. This suggests that with our approach, dialog-level feedback is almost as useful as turn-level labels. In addition, we show that using the dialogue-level reward signal for fine-tuning can further improve  supervised models in-domain.

\bibliography{domain-transfer-dst}

\begin{thebibliography}{39}
\expandafter\ifx\csname natexlab\endcsname\relax\def\natexlab#1{#1}\fi

\bibitem[{Blitzer et~al.(2006)Blitzer, McDonald, and Pereira}]{Blitzer:ea:06}
John Blitzer, Ryan McDonald, and Fernando Pereira. 2006.
\newblock {Domain adaptation with structural correspondence learning}.
\newblock In \emph{Proceedings of EMNLP}.

\bibitem[{Bowling and Veloso(2001)}]{Bowling:Veloso:01}
Michael Bowling and Manuela Veloso. 2001.
\newblock Rational and convergent learning in stochastic games.
\newblock In \emph{IJCAI}.

\bibitem[{Budzianowski et~al.(2018)Budzianowski, Wen, Tseng, Casanueva, Ultes,
  Ramadan, and Gasic}]{budzianowski2018multiwoz}
Pawe{\l} Budzianowski, Tsung-Hsien Wen, Bo-Hsiang Tseng, I{\~n}igo Casanueva,
  Stefan Ultes, Osman Ramadan, and Milica Gasic. 2018.
\newblock \href {http://aclweb.org/anthology/D18-1547} {{MultiWOZ-A Large-Scale
  Multi-Domain Wizard-of-Oz Dataset for Task-Oriented Dialogue Modelling}}.
\newblock In \emph{Proceedings of the 2018 Conference on Empirical Methods in
  Natural Language Processing}, pages 5016--5026.

\bibitem[{Chen et~al.(2018)Chen, Chang, Chen, Tan, Ga{\v{s}}i{\'c}, and
  Yu}]{chen2018policy}
Lu~Chen, Cheng Chang, Zhi Chen, Bowen Tan, Milica Ga{\v{s}}i{\'c}, and Kai Yu.
  2018.
\newblock Policy adaptation for deep reinforcement learning-based dialogue
  management.
\newblock In \emph{2018 IEEE International Conference on Acoustics, Speech and
  Signal Processing (ICASSP)}, pages 6074--6078. IEEE.

\bibitem[{{Daume III} and Marcu(2006)}]{Daume:Marcu:06}
Hal {Daume III} and Daniel Marcu. 2006.
\newblock {Domain adaptation for statistical classifiers}.
\newblock \emph{Journal of Artificial Intelligence Research}, 26:101--126.

\bibitem[{Dhingra et~al.(2017)Dhingra, Li, Li, Gao, Chen, Ahmed, and
  Deng}]{dhingra2017towards}
Bhuwan Dhingra, Lihong Li, Xiujun Li, Jianfeng Gao, Yun-Nung Chen, Faisal
  Ahmed, and Li~Deng. 2017.
\newblock Towards end-to-end reinforcement learning of dialogue agents for
  information access.
\newblock In \emph{Proceedings of the 55th Annual Meeting of the Association
  for Computational Linguistics (Volume 1: Long Papers)}, volume~1, pages
  484--495.

\bibitem[{El~Asri et~al.(2017)El~Asri, Schulz, Sharma, Zumer, Harris, Fine,
  Mehrotra, and Suleman}]{asri2017frames}
Layla El~Asri, Hannes Schulz, Shikhar Sharma, Jeremie Zumer, Justin Harris,
  Emery Fine, Rahul Mehrotra, and Kaheer Suleman. 2017.
\newblock \href {http://www.aclweb.org/anthology/W17-5526} {Frames: a corpus
  for adding memory to goal-oriented dialogue systems}.
\newblock In \emph{Proceedings of the 18th Annual SIGdial Meeting on Discourse
  and Dialogue}, pages 207--219.

\bibitem[{Gasic et~al.(2013)Gasic, Breslin, Henderson, Kim, Szummer, Thomson,
  Tsiakoulis, and Young}]{gasic2013pomdp}
Milica Gasic, Catherine Breslin, Matthew Henderson, Dongho Kim, Martin Szummer,
  Blaise Thomson, Pirros Tsiakoulis, and Steve Young. 2013.
\newblock Pomdp-based dialogue manager adaptation to extended domains.
\newblock In \emph{Proceedings of the SIGDIAL 2013 Conference}, pages 214--222.

\bibitem[{Ga{\v{s}}i{\'c} et~al.(2017)Ga{\v{s}}i{\'c}, Mrk{\v{s}}i{\'c},
  Rojas-Barahona, Su, Ultes, Vandyke, Wen, and Young}]{gavsic2017dialogue}
Milica Ga{\v{s}}i{\'c}, Nikola Mrk{\v{s}}i{\'c}, Lina~M Rojas-Barahona, Pei-Hao
  Su, Stefan Ultes, David Vandyke, Tsung-Hsien Wen, and Steve Young. 2017.
\newblock Dialogue manager domain adaptation using gaussian process
  reinforcement learning.
\newblock \emph{Computer Speech \& Language}, 45:552--569.

\bibitem[{Henderson(2015)}]{henderson2015machine}
Matthew Henderson. 2015.
\newblock Machine learning for dialog state tracking: A review.
\newblock In \emph{Proc. of The First International Workshop on Machine
  Learning in Spoken Language Processing}.

\bibitem[{Henderson et~al.(2014{\natexlab{a}})Henderson, Thomson, and
  Williams}]{henderson2014second}
Matthew Henderson, Blaise Thomson, and Jason~D Williams. 2014{\natexlab{a}}.
\newblock \href {http://www.aclweb.org/anthology/W14-4337} {The second dialog
  state tracking challenge}.
\newblock In \emph{Proceedings of the 15th Annual Meeting of the Special
  Interest Group on Discourse and Dialogue (SIGDIAL)}, pages 263--272.

\bibitem[{Henderson et~al.(2014{\natexlab{b}})Henderson, Thomson, and
  Young}]{henderson2014word}
Matthew Henderson, Blaise Thomson, and Steve Young. 2014{\natexlab{b}}.
\newblock \href {https://www.aclweb.org/anthology/W/W14/W14-4340.pdf}
  {Word-based dialog state tracking with recurrent neural networks}.
\newblock In \emph{Proceedings of the 15th Annual Meeting of the Special
  Interest Group on Discourse and Dialogue (SIGDIAL)}, pages 292--299.

\bibitem[{Jiang and Zhai(2007)}]{Jiang:Zhai:07}
Jing Jiang and ChengXiang Zhai. 2007.
\newblock {Instance weighting for domain adaptation in {NLP}}.
\newblock In \emph{Proceedings of ACL}.

\bibitem[{Kakade(2002)}]{kakade2002natural}
Sham~M Kakade. 2002.
\newblock A natural policy gradient.
\newblock In \emph{Advances in neural information processing systems}, pages
  1531--1538.

\bibitem[{Kingma and Ba(2014)}]{KingmaB14}
Diederik~P. Kingma and Jimmy Ba. 2014.
\newblock \href {http://arxiv.org/abs/1412.6980} {Adam: {A} method for
  stochastic optimization}.
\newblock \emph{ICLR}.

\bibitem[{Li et~al.(2009)Li, Williams, and
  Balakrishnan}]{Li2009ReinforcementLF}
Lihong Li, Jason~D. Williams, and Suhrid Balakrishnan. 2009.
\newblock Reinforcement learning for dialog management using least-squares
  policy iteration and fast feature selection.
\newblock In \emph{INTERSPEECH}.

\bibitem[{Liu et~al.(2018)Liu, T{\"u}r, Hakkani-T{\"u}r, Shah, and
  Heck}]{liu2018dialogue}
Bing Liu, Gokhan T{\"u}r, Dilek Hakkani-T{\"u}r, Pararth Shah, and Larry Heck.
  2018.
\newblock \href {http://aclweb.org/anthology/N18-1187} {Dialogue learning with
  human teaching and feedback in end-to-end trainable task-oriented dialogue
  systems}.
\newblock In \emph{Proceedings of the 2018 Conference of the North American
  Chapter of the Association for Computational Linguistics: Human Language
  Technologies, Volume 1 (Long Papers)}, volume~1, pages 2060--2069.

\bibitem[{Mrk{\v{s}}i{\'c} et~al.(2015)Mrk{\v{s}}i{\'c}, S{\'e}aghdha, Thomson,
  Ga{\v{s}}i{\'c}, Su, Vandyke, Wen, and Young}]{mrkvsic2015multi}
N~Mrk{\v{s}}i{\'c}, DO~S{\'e}aghdha, B~Thomson, M~Ga{\v{s}}i{\'c}, PH~Su,
  D~Vandyke, TH~Wen, and S~Young. 2015.
\newblock \href {https://aclanthology.info/papers/P15-2130/p15-2130}
  {Multi-domain dialog state tracking using recurrent neural networks}.
\newblock In \emph{ACL-IJCNLP 2015-53rd Annual Meeting of the Association for
  Computational Linguistics and the 7th International Joint Conference on
  Natural Language Processing of the Asian Federation of Natural Language
  Processing, Proceedings of the Conference}, volume~2, pages 794--799.

\bibitem[{Mrk{\v{s}}i{\'c} et~al.(2017{\natexlab{a}})Mrk{\v{s}}i{\'c},
  S{\'e}aghdha, Wen, Thomson, and Young}]{mrkvsic2017neural}
Nikola Mrk{\v{s}}i{\'c}, Diarmuid~{\'O} S{\'e}aghdha, Tsung-Hsien Wen, Blaise
  Thomson, and Steve Young. 2017{\natexlab{a}}.
\newblock \href {http://aclweb.org/anthology/P17-1163} {Neural belief tracker:
  Data-driven dialogue state tracking}.
\newblock In \emph{Proceedings of the 55th Annual Meeting of the Association
  for Computational Linguistics (Volume 1: Long Papers)}, volume~1, pages
  1777--1788.

\bibitem[{Mrk{\v{s}}i{\'{c}} and Vuli{\'{c}}(2018)}]{Mrksic:2018acl}
Nikola Mrk{\v{s}}i{\'{c}} and Ivan Vuli{\'{c}}. 2018.
\newblock \href {http://aclweb.org/anthology/P18-2018} {Fully statistical
  neural belief tracking}.
\newblock In \emph{Proceedings of ACL}, pages 108--113.

\bibitem[{Mrk{\v{s}}i{\'c} et~al.(2017{\natexlab{b}})Mrk{\v{s}}i{\'c},
  Vuli{\'c}, S{\'e}aghdha, Leviant, Reichart, Ga{\v{s}}i{\'c}, Korhonen, and
  Young}]{mrkvsic2017semantic}
Nikola Mrk{\v{s}}i{\'c}, Ivan Vuli{\'c}, Diarmuid~{\'O} S{\'e}aghdha, Ira
  Leviant, Roi Reichart, Milica Ga{\v{s}}i{\'c}, Anna Korhonen, and Steve
  Young. 2017{\natexlab{b}}.
\newblock \href {http://www.aclweb.org/anthology/Q17-1022} {Semantic
  specialization of distributional word vector spaces using monolingual and
  cross-lingual constraints}.
\newblock \emph{Transactions of the Association of Computational Linguistics},
  5(1):309--324.

\bibitem[{Nouri and Hosseini-Asl(2018)}]{nouri2018gce}
Elnaz Nouri and Ehsan Hosseini-Asl. 2018.
\newblock Toward scalable neural dialogue state tracking.
\newblock In \emph{NeurIPS 2018, 2nd Conversational AI workshop}.

\bibitem[{Papini et~al.(2017)Papini, Pirotta, and
  Restelli}]{papini2017adaptive}
Matteo Papini, Matteo Pirotta, and Marcello Restelli. 2017.
\newblock Adaptive batch size for safe policy gradients.
\newblock In \emph{Advances in Neural Information Processing Systems}, pages
  3591--3600.

\bibitem[{Paszke et~al.(2017)Paszke, Gross, Chintala, Chanan, Yang, DeVito,
  Lin, Desmaison, Antiga, and Lerer}]{paszke2017automatic}
Adam Paszke, Sam Gross, Soumith Chintala, Gregory Chanan, Edward Yang, Zachary
  DeVito, Zeming Lin, Alban Desmaison, Luca Antiga, and Adam Lerer. 2017.
\newblock \href {https://openreview.net/pdf?id=BJJsrmfCZ} {Automatic
  differentiation in pytorch}.
\newblock In \emph{NIPS-W}.

\bibitem[{Ramadan et~al.(2018)Ramadan, Budzianowski, and
  Gasic}]{ramadan2018large}
Osman Ramadan, Pawe{\l} Budzianowski, and Milica Gasic. 2018.
\newblock \href {http://www.aclweb.org/anthology/P18-2069} {Large-scale
  multi-domain belief tracking with knowledge sharing}.
\newblock In \emph{Proceedings of the 56th Annual Meeting of the Association
  for Computational Linguistics (Volume 2: Short Papers)}, volume~2, pages
  432--437.

\bibitem[{Rastogi et~al.(2017)Rastogi, Hakkani-T{\"u}r, and
  Heck}]{rastogi2017scalable}
Abhinav Rastogi, Dilek Hakkani-T{\"u}r, and Larry Heck. 2017.
\newblock \href
  {https://static.googleusercontent.com/media/research.google.com/en//pubs/archive/46399.pdf}
  {Scalable multi-domain dialogue state tracking}.
\newblock In \emph{2017 IEEE Automatic Speech Recognition and Understanding
  Workshop (ASRU)}, pages 561--568. IEEE.

\bibitem[{Ren et~al.(2018)Ren, Xie, Chen, and Yu}]{ren2018towards}
Liliang Ren, Kaige Xie, Lu~Chen, and Kai Yu. 2018.
\newblock \href {http://aclweb.org/anthology/D18-1299} {Towards universal
  dialogue state tracking}.
\newblock In \emph{Proceedings of the 2018 Conference on Empirical Methods in
  Natural Language Processing}, pages 2780--2786.

\bibitem[{Singh et~al.(2002)Singh, Litman, Kearns, and
  Walker}]{singh2002optimizing}
Satinder Singh, Diane Litman, Michael Kearns, and Marilyn Walker. 2002.
\newblock \href {https://dl.acm.org/citation.cfm?id=1622410} {Optimizing
  dialogue management with reinforcement learning: Experiments with the njfun
  system}.
\newblock \emph{Journal of Artificial Intelligence Research}, 16:105--133.

\bibitem[{Sun et~al.(2014)Sun, Chen, Zhu, and Yu}]{sun2014generalized}
Kai Sun, Lu~Chen, Su~Zhu, and Kai Yu. 2014.
\newblock A generalized rule based tracker for dialogue state tracking.
\newblock In \emph{2014 IEEE Spoken Language Technology Workshop (SLT)}, pages
  330--335. IEEE.

\bibitem[{Sun et~al.(2016)Sun, Zhu, Chen, Yao, Wu, and Yu}]{sun2016hybrid}
Kai Sun, Su~Zhu, Lu~Chen, Siqiu Yao, Xueyang Wu, and Kai Yu. 2016.
\newblock Hybrid dialogue state tracking for real world human-to-human
  dialogues.
\newblock In \emph{INTERSPEECH}, pages 2060--2064.

\bibitem[{Sutton and Barto(1998)}]{Sutton:1998}
Richard~S. Sutton and Andrew~G. Barto. 1998.
\newblock \href {http://www.incompleteideas.net/book/first/the-book.html}
  {\emph{Introduction to Reinforcement Learning}}, 1st edition.
\newblock MIT Press, Cambridge, MA, USA.

\bibitem[{Wang and Lemon(2013)}]{wang2013simple}
Zhuoran Wang and Oliver Lemon. 2013.
\newblock \href {https://www.aclweb.org/anthology/W/W13/W13-4067.pdf} {A simple
  and generic belief tracking mechanism for the dialog state tracking
  challenge: On the believability of observed information}.
\newblock In \emph{Proceedings of the SIGDIAL 2013 Conference}, pages 423--432.

\bibitem[{Weaver and Tao(2001)}]{Weaver:Tao:01}
Lex Weaver and Nigel Tao. 2001.
\newblock The optimal reward baseline for gradient-based reinforcement
  learning.
\newblock In \emph{UAI}.

\bibitem[{Williams(2013)}]{W13-4068}
Jason Williams. 2013.
\newblock \href {http://aclweb.org/anthology/W13-4068} {Multi-domain learning
  and generalization in dialog state tracking}.
\newblock In \emph{Proceedings of the SIGDIAL 2013 Conference}, pages 433--441.
  Association for Computational Linguistics.

\bibitem[{Williams et~al.(2013)Williams, Raux, Ramachandran, and
  Black}]{williams2013dialog}
Jason Williams, Antoine Raux, Deepak Ramachandran, and Alan Black. 2013.
\newblock \href {http://www.aclweb.org/anthology/W13-4065} {The dialog state
  tracking challenge}.
\newblock In \emph{Proceedings of the SIGDIAL 2013 Conference}, pages 404--413.

\bibitem[{Williams and Young(2007)}]{WILLIAMS2007393}
Jason~D. Williams and Steve Young. 2007.
\newblock \href {https://doi.org/https://doi.org/10.1016/j.csl.2006.06.008}
  {Partially observable markov decision processes for spoken dialog systems}.
\newblock \emph{Computer Speech \& Language}, 21(2):393 -- 422.

\bibitem[{Williams and Peng(1991)}]{Williams1991FunctionOU}
Ronald~J Williams and Jing Peng. 1991.
\newblock Function optimization using connectionist reinforcement learning
  algorithms.
\newblock \emph{Connection Science}, 3(3):241--268.

\bibitem[{Zhao and Eskenazi(2016)}]{zhao2016}
Tiancheng Zhao and Maxine Eskenazi. 2016.
\newblock \href {https://doi.org/10.18653/v1/W16-3601} {Towards end-to-end
  learning for dialog state tracking and management using deep reinforcement
  learning}.
\newblock pages 1--10.

\bibitem[{Zhong et~al.(2018)Zhong, Xiong, and Socher}]{zhong2018global}
Victor Zhong, Caiming Xiong, and Richard Socher. 2018.
\newblock \href {http://www.aclweb.org/anthology/P18-1135} {Global-locally
  self-attentive encoder for dialogue state tracking}.
\newblock In \emph{Proceedings of the 56th Annual Meeting of the Association
  for Computational Linguistics (Volume 1: Long Papers)}, volume~1, pages
  1458--1467.

\end{thebibliography}
\bibliographystyle{acl_natbib}

\appendix
\newpage
\section{Appendix}
\label{sec:appendix}

\begin{table*}[h!]
    \scriptsize
    \centering
    \begin{tabular}{cccc}
    \toprule
        \textbf{System utterance} & \textbf{User utterance} & \textbf{Baseline prediction} & \textbf{PG fine-tune prediction}  \\
        \midrule
    N/A   & \makecell{I'm looking for a cheap\\ place to dine, preferably \\in the centre of town.} & \makecell[l]{\textcolor{OliveGreen}{inform(area=center)}\\\textcolor{red}{inform(pricerange=expensive)}} & \makecell[l]{\textcolor{OliveGreen}{inform(area=center)}\\\textcolor{OliveGreen}{inform(pricerange=cheap)}} \\
        \midrule
\makecell{Yes, I have 4 results matching\\ your request, is there a price \\ range you're looking for?}   & \makecell{I would like moderate\\ price range please.} & \makecell[l]{\textcolor{red}{inform(pricerange=expensive)}} & \makecell[l]{\textcolor{OliveGreen}{inform(pricerange=moderate)}} \\
        \midrule
\makecell{There are a number of options \\ for Indian restaurants in the \\ centre of town.What \\ price range would you like ?}   & \makecell{I would prefer cheap \\ restaurants.} & \makecell[l]{\textcolor{red}{inform(pricerange=expensive)}} & \makecell[l]{\textcolor{OliveGreen}{inform(pricerange=cheap)}} \\
\bottomrule
    \end{tabular}
    \caption{Comparison of example turn predictions from the MultiWOZ dataset between the baseline model trained on the \textsc{hotel} domains, and the policy gradient fine-tuned model. Green indicates a correct prediction whereas red indicates a wrong prediction.}
    \label{tab:examples}
\end{table*}


\end{document}